%% file: ms.tex
\documentclass{article}

\usepackage{arxiv}
\usepackage{hyperref}       
\usepackage{url}            
\usepackage{booktabs}       
\usepackage{amsfonts}       
\usepackage{nicefrac}       
\usepackage{microtype}      
\usepackage{lipsum}
\usepackage{graphicx}
\usepackage{xcolor}
\usepackage{caption}
\usepackage{multirow}
\usepackage{amsmath}
\usepackage[export]{adjustbox}

\hypersetup{
pdftitle={Bayesian Hierarchical Models for Counterfactual Estimation},
pdfauthor={Natraj Raman}
}

\title{Bayesian Hierarchical Models for Counterfactual Estimation}

\author{
 Natraj Raman$^{1}$, Daniele Magazzeni$^{1}$ and Sameena Shah$^{2}$ \\
  JPMorgan AI Research\\
  $^{1}$London, UK. \\
  $^{2}$New York, USA. \\
  \texttt{first.last@jpmorgan.com} \\
}

\begin{document}
\maketitle
\input{sec_abstract}

\input{sec_intro}
\input{sec_relwork}
\input{sec_model}
\input{sec_results}
\input{sec_conclusion}

\section*{Acknowledgments}{This paper was prepared for information purposes by the Artificial Intelligence Research group of JPMorgan Chase \& Co and its affiliates (“JP Morgan”), and is not a product of the Research Department of JP Morgan.  J.P. Morgan makes no representation and warranty whatsoever and disclaims all liability for the completeness, accuracy or reliability of the information contained herein. This document is not intended as investment research or investment advice, or a recommendation, offer or solicitation for the purchase or sale of any security, financial instrument, financial product or service, or to be used in any way for evaluating the merits of participating in any transaction, and shall not constitute a solicitation under any jurisdiction or to any person, if such solicitation under such jurisdiction or to such person would be unlawful. © 2021 JP Morgan Chase \& Co. All rights reserved.}

\bibliographystyle{unsrt}  
\bibliography{references}  

\section*{Appendix}
\appendix

\input{supplement}

\end{document}

%% file: sec_abstract.tex
\begin{abstract}
Counterfactual explanations utilize feature perturbations to analyze the outcome of an original decision and recommend an actionable recourse. We argue that it is beneficial to provide several alternative explanations rather than a single point solution and propose a probabilistic paradigm to estimate a diverse set of counterfactuals. Specifically, we treat the perturbations as random variables endowed with prior distribution functions. This allows sampling multiple counterfactuals from the posterior density, with the added benefit of incorporating inductive biases, preserving domain specific constraints and quantifying uncertainty in estimates. More importantly, we leverage Bayesian hierarchical modeling to share information across different subgroups of a population, which can both improve robustness and measure fairness. A gradient based sampler with superior convergence characteristics efficiently computes the posterior samples. Experiments across several datasets demonstrate that the counterfactuals estimated using our approach are valid, sparse, diverse and feasible.
\end{abstract}

\keywords{Diverse Counterfactuals \and Personalized Recourse \and Hierarchical Bayes \and Hamiltonian Monte Carlo}

%% file: sec_intro.tex
\section{Introduction}
Large-scale adoption of decision critical AI solutions requires explaining the rationale for a particular prediction. Counterfactual explanations~\cite{wachter2017counterfactual} provide an intuitive mechanism to reason even over complex models by defining a set of feature perturbations that would change the outcome to a favourable decision, thereby explaining the factors that led to the original decision.

Many counterfactuals are possible since there are several paths for an instance to achieve the desired outcome. After all, a loan applicant could qualify for a mortgage with different combinations of increasing the collateral amount, obtaining an educational degree or working longer hours. The focus traditionally~\cite{dhurandhar2018explanations, pawelczyk2020learning, looveren2021interpretable} though has been on finding a unique path that is defined by the smallest possible perturbation of an instance.   We argue that a theoretically ideal single explanation is overly restrictive and propose a mechanism to offer multiple alternative counterfactual explanations in a Bayesian framework. 

While there has been some investigations before ~\cite{rodriguez2021beyond, mothilal2020explaining, russell2019efficient} on generating diverse counterfactuals, they have largely been cast in the frequentist optimization setting producing point-estimates. In contrast, we follow a Bayesian approach that models the perturbations in a probabilistic paradigm and regard them as random variables with distribution functions. 

\begin{figure*}[!htbp]
\centering
\includegraphics[width=\linewidth]{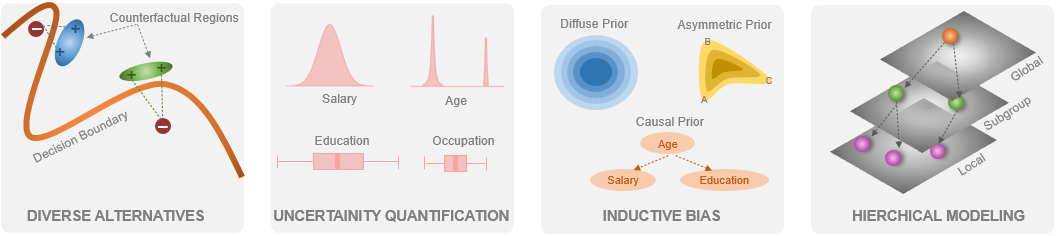}
\caption{Advantages of Bayesian approach to counterfactual estimation. Several alternative explanations can be sampled from the posterior distribution, the uncertainty in predicted variables can be summarized through credible intervals, user specific customization can be formalized using informative priors and the hierarchical structure can capture relative effects.  } 
\label{fig_overview}
\end{figure*}

This probabilistic treatment offers a variety of benefits over traditional models as summarized in Figure ~\ref{fig_overview}. For example, the posterior distribution of the perturbations can be directly used to sample a diverse set of counterfactuals in a single shot without resorting to the use of ensemble models or exotic post-hoc selection constraints. A distribution oriented approach~\cite{gelman1995bayesian} is asymptotically unbiased, can natively handle multimodal parameters and quantify the uncertainty in predictions using variances, interquartile ranges, credible intervals and entropy. Such variability measures can help in producing robust counterfactual estimates, thereby improving the overall reliability. 

Another appealing aspect of our approach is the ability to customize the counterfactual generation through appropriate prior distributions. Consider a user who is less likely to take a \emph{Sales} job when compared with other job types. We can encode this belief through an asymmetric categorical prior that places reduced probability mass on the \emph{Sales} type, and vary this mass depending on the relative extent of the user's preference. The incorporation of such specific knowledge through informative priors enables greater flexibility and personalization in counterfactual generation. 

The generated counterfactuals must respect the inherent dependencies between the features in order to produce examples that are feasible~\cite{mahajan2019preserving, poyiadzi2020face} in the real world. For example, we may want to preserve the correlation between \emph{Age} and \emph{Salary} variables in the observed data. Our solution allows the introduction of conditional distributions that can capture the causal structure relating the variables, thereby modeling the feature interactions explicitly. 

When presenting a counterfactual, it is useful to compare the magnitude of change in the feature value of an instance with population level counterfactuals. For example, in the spirit of fairness~\cite{wu2019counterfactual}, it may be of interest to determine whether the education level suggested by the counterfactual to get a loan is discriminative or not relative to other typical instances. These comparisons could also span across multiple hierarchical levels considering demographic subgroups within the population such as \emph{Male} and \emph{Female} or \emph{White}, \emph{Black} and \emph{Asian}.  A natural way to model these dependencies across instances is by using a Bayesian hierarchical structure that can share information across data groups and yet capture variations at different levels of the data, thereby improving the quality of the estimates and enabling fairness evaluation.

We propose here a three-level hierarchical counterfactual model in which the perturbations of a local instance depends on its corresponding subgroup, which in turn depends on the population. We account both for continuous and categorical variables, treating the latter as first-class citizens. Due to the intractable nature of the posterior distribution, we utilize Hamiltonian Monte Carlo (HMC)~\cite{betancourt2015hamiltonian} to derive the posterior samples. HMC makes use of the geometric information estimated via first-order derivatives of the target distribution and this enables us to efficiently explore the parameter space. 

We use three public tabular datasets namely Adult Income ~\cite{adult1996}, German Credit ~\cite{german1994} and HELOC ~\cite{heloc2018} for demonstrating the efficacy of our approach. In particular, we show that the Bayesian model compares well with point-estimate based solutions in measures such as validity, proximity and sparsity, with the added benefit of its support for diversity, robustness, causality, personalization and borrowing strength.

To summarize, our main contributions are: (a) A distribution oriented approach to counterfactual generation that can produce diverse alternatives and quantify uncertainty, (b) Incorporation of inductive biases through informative priors, and (c) A hierarchical formulation that preserves feature dependencies, promotes information sharing and enables subgroup analysis.

%% file: sec_relwork.tex
\section{Related Work}\label{sec_relwork}
Counterfactual explanations has several research themes~\cite{verma2020counterfactual} and our primary focus is on the diversity angle. While most methods produce a single counterfactual~\cite{dhurandhar2018explanations, pawelczyk2020learning, looveren2021interpretable}, the generation of multiple counterfactuals had been considered before.  The preferred route to support alternate examples is to incorporate an explicit term in the optimization objective function or perform an efficient search. For example in ~\cite{mothilal2020explaining}, a diversity constraint is included in the loss function by building on detrimental point processes, while in ~\cite{rodriguez2021beyond}, a collection of latent perturbations are searched to identify the attributes that will change the decision. ~\cite{russell2019efficient} employs a mixed-integer programming solver that is integrated with a set of constraints to generate diverse explanations. More recently, ~\cite{smyth2022few} follow an endogenous approach in which the nearest neighbours of feature values are searched to retrieve multiple counterfactuals. 

The primary difference with the above methods is that we cast counterfactual generation as sampling from posterior distributions. Even methods that claim to be distribution-aware deal with the empirical data distribution~\cite{kanamori2020dace} or the distribution of classifier model parameters~\cite{bui2021counterfactual}, unlike our focus on the counterfactual generation process. 

A distribution-centric framework such as ours produces uncertainty estimates that can be used to measure the stability, reliability and consistency of the generated counterfactuals. Few recent works such as the extensions to LIME and SHAP in ~\cite{slack2021reliable}, the latent variable deep generative model in ~\cite{ley2022diverse} and the optimal transport for Gaussian mixtures in ~\cite{nguyen2022robust} do consider uncertainty estimates. However, their focus is either on generating local explanations of a classifier prediction or on the predictive uncertainty of the classification model.

A further distinguishing aspect of our work is the treatment of counterfactuals in a hierarchical setting. Existing methods that can provide explanations at local and global levels such as ~\cite{plumb2020explaining, becker2021step} lack a principled framework that can inherently support multiple levels as ours does. Works that can potentially support several levels, as in ~\cite{rawal2020beyond, kanamori2022counterfactual}, differ from ours in their objective and technique.

%% file: sec_model.tex
\section{Model}\label{sec_model}
Let $\mathbf{x}=(x_1,...,x_{d_{cont}},x_{d_{cont}+1},...,x_{d_{cont}+d_{cat}})$ be an observed instance that contains $d_{cont}$ number of continuous values and $d_{cat}$ number of categorical values with $|\mathbf{x}|=d_{cont}+d_{cat}=d$. Each instance is associated with one of $K$ different groups and let $k$ be the subgroup corresponding to $\mathbf{x}$. Let $f: x \to [0,1]$ be a binary classifier function that is differentiable. 

If $f(\mathbf{x}) <= 0.5$, we would like to generate a counterfactual explanation $\mathbf{x}^{*}$ for $\mathbf{x}$ such that $f(\mathbf{x}^{*}) > 0.5$. We write 
\begin{align}
\begin{split}
\mathbf{x}^{*} & =  \mathbf{z}\odot\mathbf{x} + \Delta, \\  
\mathbf{\Delta} & = (\delta_1,...,\delta_{d_{cont}},\eta_{d_{cont}+1},...,\eta_{d}),
\end{split}
\end{align}
where $\Delta$ is a set of parameters that models the perturbations, with $\delta$ being the change in value of continuous variables and $\eta$ the modified value of a categorical variable. Here $\mathbf{z} \in \{0,1\}^d$ is a vector of indicators that is set to 1 for continuous variables and $\odot$ denotes element-wise multiplication.

In traditional counterfactual frameworks, $\Delta$ is a set of fixed but unknown values that must be determined based on some optimization function. Instead, in our Bayesian paradigm, $\Delta$ is modeled as a random variable having a probability distribution. This allows the incorporation of problem specific knowledge (or the lack of it) on these parameters through appropriate prior distributions. Furthermore, we can sample multiple values corresponding to different modes of the distribution, thereby generating a diverse set of explanations.

\subsection{Hierarchical Bayes}
The modular nature of a Bayesian framework allows simplifying complex setups necessitated by counterfactual generation requirements. For example, we may wish to measure how a counterfactual generated for a particular instance differs from other instances in its subgroup or across the groups. This requirement to capture variations at different levels of the data can be formulated naturally as a hierarchical model, where we have a population level set of parameters that are shared with the parameters at a group level which in turn are shared at the individual instance level. Such a hierarchical structure also enables borrowing of information from other related data, through which the robustness of parameter estimates can be improved.

\begin{figure}[!tbp]
\centering
\includegraphics{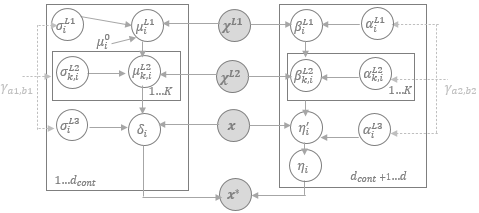}
\caption{Hierarchical counterfactual model in plate notation. The left side of the figure corresponds to the dependency graph of continuous variables while the right side reflects the categorical variables.  } 
\label{fig_model}
\end{figure}

We describe a three level hierarchical counterfactual model with the population level being referred as $L1$, the group level as $L2$ and the local instance level as $L3$. Each continuous variable $\delta$ is bestowed a Gaussian prior $\mathcal{N}(\mu,\sigma)$, with the mean $\mu$ being shared across the hierarchical levels while the variance $\sigma$ remains independent. Intuitively, the parameter value of a subgroup is centered around the population level value but can still deviate from it. Similarly, the parameters of all the local instances can take distinct values, yet are dependent on its parent and remains within a range of its corresponding subgroup. A level can optionally be dropped, in which case the dependency shifts upwards. Formally, the generative steps are defined as:
\begin{align}
\mu^{L1}_{i} &\sim \mathcal{N}(\mu^{0}_{i}, \sigma^{L1}_i)  & \forall i &= 1...d_{cont}\\
\mu^{L2}_{k,i} &\sim \mathcal{N}(\mu^{L1}_{i}, \sigma^{L2}_{k,i}) & \forall i &= 1...d_{cont}, k=1...K\\
\delta_i &\sim \mathcal{N}(\mu^{L2}_{k,i}, \sigma^{L3}_i) & \forall i &= 1...d_{cont}.
\end{align}

The relationship between the variables are modeled using conditional distributions. Given a causal model where variable $i$ is the parent of $j$, the dependency structure is defined using a linear approximation with parameters $m$ and $c$ as
\begin{align}
\delta_j | \delta_i &\sim \mathcal{N}(m\delta_i+c, \sigma_j).
\end{align}

A categorical variable $\eta$ is endowed with a scaled Dirichlet prior $Dir(\alpha\beta)$, where $\beta \in \mathbb{R}_{+}^{L}$ is a vector whose length depends on the number of categories $L$ and $\alpha$ is a scale parameter. The categorical values themselves are sampled from the probabilities estimated from the Dirichlet prior using a Multinomial distribution.  Similar to the continuous variables, each parameter depends on the parameter at its corresponding parent level while still retaining the flexibility to diverge. The generative steps here are formalized as:
\begin{align}
\beta^{L1}_{i} &\sim Dir(\alpha^{L1}_i) & \forall i &= d_{cont}\text{\footnotesize $+1$}...d\\
\beta^{L2}_{k,i} &\sim Dir(\alpha^{L2}_{k,i}\beta^{L1}_{i}) & \forall i &= d_{cont}\text{\footnotesize $+1$}...d, k=1...K\\
\eta^{'}_i &\sim Dir(\alpha^{L3}_i\beta^{L2}_{k,i}) & \forall i &= d_{cont}\text{\footnotesize $+1$}...d\\ 
\eta_i &\sim Mult(\eta^{'}_i) & \forall i &= d_{cont}\text{\footnotesize $+1$}...d.
\end{align}

The variance parameters $\sigma$ are further assigned independent inverse gamma priors with hyper-parameters $(\gamma_{a1},\gamma_{b1})$. Similarly, $\alpha$ follows a gamma prior with hyper-parameters $(\gamma_{a2},\gamma_{b2})$. The conditional dependency graph of the model is illustrated in Figure ~\ref{fig_model}. 

\subsection{Posterior Inference}
Let $\Theta=\{\Delta,\mu,\sigma,\alpha,\beta\}$ be the set of all parameters. The central inference problem is to estimate these parameters given an observed instance $\mathbf{x}$ and a training dataset 
\begin{align}
\mathcal{D}=\text{\small $( \chi^{L2+}_1, ..., \chi^{L2+}_K, \chi^{L2-}_1, ..., \chi^{L2-}_K, \chi^{L1+}, \chi^{L1-})$},
\end{align}
where $\chi^{L2+}_k$ represents a set of training instances corresponding to subgroup $k$ such that $f(.) > 0.5$ while $\chi^{L1+}$ is the set of instances pooled across all the subgroups. Similarly $\chi^{L2-}_k$ and  $\chi^{L1-}$ denote the negative instances. The posterior density of the parameters is given as
\begin{align}
p(\Theta|\mathbf{x},\mathcal{D}) = \int & f(\mathbf{x}^*|\Theta) p(\mathbf{x}|\mathbf{x}^*,\Theta) \nonumber \\ & \prod_k{p(\chi^{L2}_k|\Theta)}p(\chi^{L1}|\Theta) p(\Theta) d\Theta.
\end{align}
Here $\mathbf{x}^*$ is constructed as in equation (1) and $p(\Theta)$ is the prior distribution for the parameters as defined in equations (2) to (9). The first likelihood term in the above posterior captures the probability of a counterfactual to be valid while the second term encourages close proximity between an original instance and its counterfactual. Specifically,
\begin{align}
p(\mathbf{x}|\mathbf{x}^*,\Theta) \propto e^{\text{\large $\frac{-\|\mathbf{x}^*-\mathbf{x}\|_2}{2\lambda}$}},
\end{align}
where $\lambda$ is a bandwidth hyper-parameter. 

The likelihood of the training instances at the population level is written as
\begin{align}
p(\chi^{L1}|\Theta) \propto \prod_{\mathbf{y} \in \chi^{L1-}}{f(\mathbf{y}^*|\Theta)}  e^{\text{\large $\frac{-\|\mathbf{y}^*-\bar{\chi}^{L1+}\|_2}{2\lambda}$}}
\end{align}
where $\mathbf{y^*}$ is the counterfactual constructed from a negative training instance based on the parameters at $L1$ and  $\bar{\chi}^{L1+}$ is the expected value of the positive instances in the feature space. Intuitively, we wish to find the counterfactuals of negative training instances at the population level in such a way that they are in close proximity to the positive instances of the training data. A similar construct follows for $p(\chi^{L2}_k|\Theta)$.

\subsection {Sampling Mechanism}
An approximation to equation (11) must be developed since exact posterior inference is not feasible. Simulation techniques such as Markov Chain Monte Carlo (MCMC)~\cite{neal1993probabilistic} methods allow drawing a sequence of correlated samples that can be used to estimate the intractable integrals. However, considering the large number of parameters and the complex nature of the posterior distribution induced by the presence of the classifier function, traditional MCMC methods such as Metropolis and Gibbs sampling will struggle to converge to the target distribution.

Hamiltonian Monte Carlo (HMC)~\cite{betancourt2015hamiltonian} sampling methods enable efficient exploration of such complex parameter spaces by incorporating the gradient of the log posterior. The geometric information provided by these gradients can guide the chain towards regions of high posterior density, thereby reducing the number of samples required for convergence. Exploiting the fact that $f(.)$ is differentiable, we use the No-U-Turn-Sampler (NUTS)~\cite{hoffman2014no} variant of HMC  for computing the posterior samples. 

The $(\Delta_1,...\Delta_N)$ samples produced by this sampling mechanism are used to derive $N$ different ${\mathbf{x}^*}$ that can serve as a diverse set of counterfactuals for a given $\mathbf{x}$. The uncertainty in the generated samples can be quantified using measures such as variances, interquartile ranges and credible intervals, while a point-estimate if required can be obtained through summaries or ranking the samples by cost metrics.

%% file: sec_results.tex
\section{Experiments}\label{sec_results}
We discuss here the experiment setup, present counterfactual evaluations and assess convergence properties. More details can be found in the supplementary material.

\subsection{Evaluation Setup}
\noindent \textbf{Datasets}: 
To evaluate our approach, we consider the following datasets: (a) Adult Income ~\cite{adult1996} - a dataset containing the income factors of various individuals such as \emph{Gender}, \emph{Race}, \emph{Marital Status}, \emph{Education}, \emph{Workclass}, \emph{Occupation}, \emph{Age} and \emph{Hours}. Except for \emph{Age} and \emph{Hours}, the rest of the variables are categorical and the classification objective is to predict whether an individuals' income exceeds \$50K.  (b) German Credit ~\cite{german1994} - a dataset that includes 20 different attributes of persons who takes a credit in a bank, with the vast majority of these attributes being categorical and a binary label indicating if an individual is a credit risk or not. and (c) HELOC ~\cite{heloc2018} - a dataset with information on customers who received a home equity line of credit. It has over over 20 features that are predominantly continuous and a binary label as to whether a customer paid back the loan or not. 

\noindent \textbf{Classification Model}: 
We train a non-linear neural-network model with 2 layers and 200 hidden neurons as the classifier. The categorical values are converted to a smoothed one-hot encoded vector while the continuous values are normalized to be between 0 and 1 in the feature space. We obtain an accuracy of 80\% for Adult Income, 75\% for German Credit and 72\% for HELOC. We use only the instances that are correctly classified by the model in ground-truth when evaluating the counterfactuals.

\noindent \textbf{Settings}: 
Throughout the experiments, we used a burn-in of $5000$ samples, and a target sample size of $1000$, which is sufficiently large. The standard normal distribution and a symmetric Dirichlet was used. The gamma hyper-parameters were set to a unit value and a value of 0.7 was used for the bandwidth.

\subsection{Flattened Bayes Evaluation}
We first focus on the assessment of a flattened model where only the local instances are considered. This would allow comparing the benefits of using a Bayesian model over a traditional point-estimate based counterfactual model.

\begin{table}[!tbp]
\caption{Examples of generated counterfactual samples. First row is an original instance in the Adult Income dataset. }
\centering
\small
\begin{tabular}{|l|l|l|l|l|}
\hline
  Workclass  & Education & Occupation & Hours & Label \\
  \hline

Self-emp & HS-grad & Sales & 30 & \textless=50K \\
\hline
Self-emp & Prof-School & Sales & 30 & \textgreater50K \\
Self-emp & HS-grad & Blue-Collar & 43 & \textgreater50K \\
Gov & HS-grad & Professional & 42 & \textgreater50K \\
Private & Masters & White-Collar & 25 & \textgreater50K \\
\hline
\end{tabular}
\label{tab_samplecfe}
\end{table}

\begin{table*}[!tbp]
\caption{Comparison of generated counterfactuals between Bayesian Posteriors and Point Estimates. }
\centering
\begin{tabular}{|l|r|r|r|r|r|r|}
\hline
 \multicolumn{1}{|l|}{}&\multicolumn{3}{c|}{Bayesian Posterior}&\multicolumn{3}{c|}{Point Estimates} \\
 \cline{2-4} \cline{5-7}
 Dataset & Validity \%  & Sparsity \% & Proximity  & Validity \%  & Sparsity \% & Proximity  \\
  \hline

Adult Income & 100 & 50 & 1.8  & 100 & 38 & 1.2 \\
German Credit & 97 & 50 & 2.8  & 100 & 35 & 2.5 \\
HELOC & 94 & 48 & 3.3  & 91 & 36 & 2.3 \\

\hline
\end{tabular}
\label{tab_cfecomp}
\end{table*}

\noindent \textbf{Qualitative}: Table \ref{tab_samplecfe} shows a few examples of counterfactual samples that were generated for an instance in Adult Income dataset, where a realistic setting is followed in which variables such as \emph{Age}, \emph{Gender}, \emph{Race} and \emph{Marital Status} are frozen while the values of other variables are allowed to change. The generated samples provide a wider range of options to the target user and highlights the various possibilities to have an income above \$50K. For example, the first sample indicates an option where the user can continue to retain the current values of \emph{Workclass}, \emph{Occupation} and \emph{Hours}, while the second sample allows to retain only \emph{Workclass} and \emph{Education}. The samples also help address questions such as \emph{"What if I have to work fewer hours than I currently do"}.  Contrast this with a point-estimate model where the counterfactual must be re-generated for each \emph{what-if} question and the potential set of options for the user may not be evident. 

\noindent \textbf{Quantitative}: We also perform a quantitative comparison of the Bayesian method with the traditional point-estimate technique over various measures such as validity, sparsity and proximity for all the three datasets in Table \ref{tab_cfecomp}. All the features are assumed to be mutable here. The validity metric captures the percentage of negative instances in the test set for which a valid counterfactual was generated, and the coverage appears nearly identical between the two methods. The sparsity metric highlights the percentage of features being used, while proximity is a measure of the closeness between the counterfactual and the original instance in the feature space. The point-estimates seem to perform slightly better for these two attributes. This is unsurprising because the variety in samples is an important consideration of the Bayesian method and this requires an increase both in the number and change in magnitude of the features.

\begin{figure*}[!tbp]
\centering
\includegraphics[width=\textwidth]{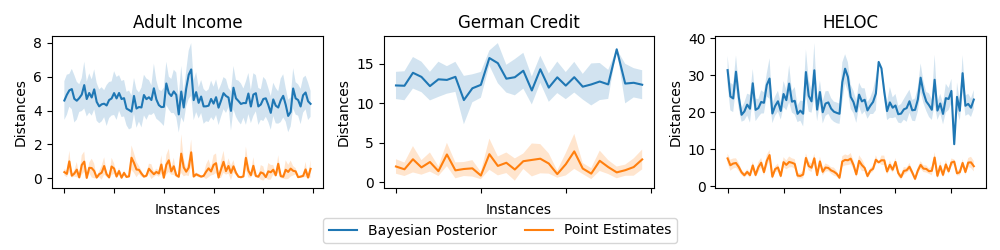}
\caption{Diversity comparison between Bayesian Posteriors and randomly initialized Point-Estimates for different datasets. Larger distances imply greater diversity between the counterfactual samples of an instance.} 
\label{fig_divcomp}
\end{figure*}

\begin{figure*}[!tbp]
\centering
\includegraphics[width=\textwidth]{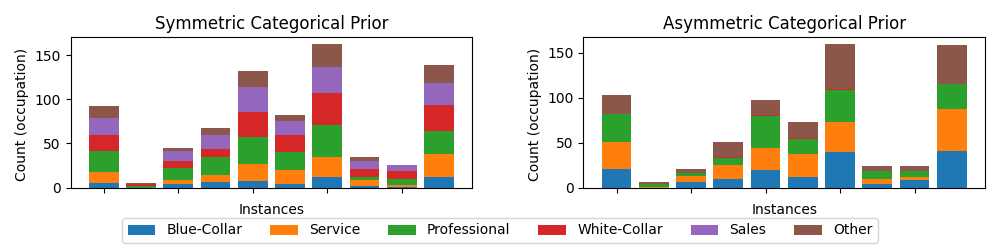}
\caption{Inductive bias in categorical values of generated counterfactuals. \emph{left}: Posterior samples from a symmetric Dirichlet prior on the \emph{Occupation} categorical variable. \emph{right}: Asymmetric prior with negligible mass on \emph{Sales} and \emph{White-Collar} values.} 
\label{fig_priorimpact}
\end{figure*}

\begin{figure}[!tbp]
\centering
\includegraphics[height=4cm]{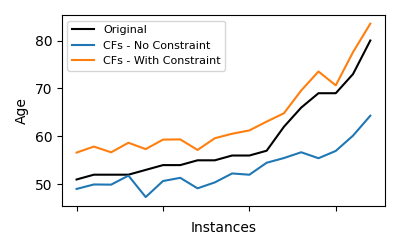}
\caption{Unary constraint ensuring that the generated counterfactuals increase over the original.} 
\label{fig_constrposage}
\end{figure}

\begin{figure}[!tbp]
\centering
\includegraphics[height=4cm]{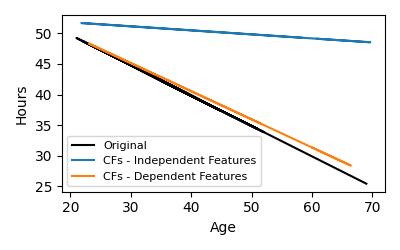}
\caption{Binary constraint preserving the feature relationship between a variable pair.} 
\label{fig_constragehours}
\end{figure}

\begin{figure*}[!tbp]
\centering
\includegraphics[width=\textwidth]{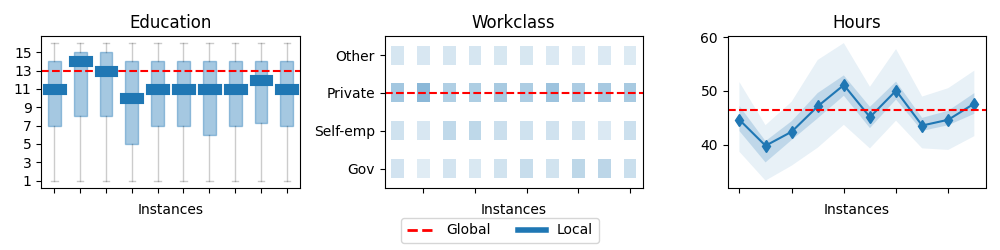}
\caption{Global vs Local Counterfactuals in Adult Income. \emph{left}: Box plot of ordinal variable \emph{Education} with the median value of samples for different instances mostly below the global value 13. \emph{middle}: The samples mode coincides with the global for nominal variable \emph{Workclass}. \emph{right}: The credible interval region for continuous variable \emph{Hours} covers the global value.}
\label{fig_hier1_ad}
\end{figure*}

\noindent \textbf{Diversity}: 
A key benefit of the Bayesian counterfactual method is its ability to generate multiple options for the user. It is desirable for the counterfactual samples not to be trivial modifications and rather be a  diverse choice. We compare the diversity of the counterfactuals produced by the Bayesian model with a point-estimate model that is initialized with different random seeds. The diversity metric is computed by calculating the distance between a counterfactual and all its peer samples and averaging over them. For a continuous variable the distance is based on the $l1$ norm scaled by median absolute deviation in the training set, while for a  categorical variable it is based on whether the value has changed or not. Figure ~\ref{fig_divcomp} illustrates this comparison by plotting the mean distance and their variance. The optimization function used in point-estimate models seem to converge towards a narrow region despite random initialization. In contrast, it is evident that across datasets the Bayesian methods owing to their distribution oriented approach produce considerably diverse samples. 

\noindent \textbf{Inductive Bias}: When generating the counterfactuals for an instance, we may want to incorporate customizations that are driven by apriori beliefs. For instance, a user may be less inclined to change their education levels beyond a Masters degree or would prefer not working longer hours. Such beliefs can be naturally inducted through asymmetric and truncated Bayesian priors. Figure ~\ref{fig_priorimpact} presents the samples generated from two different priors for the categorical variable \emph{Occupation} in Adult Income dataset. The left side figure uses a symmetric Dirichlet prior with all the categories being equally likely. The right side figure uses an asymmetric prior that reduces the mass on \emph{White-Collar} category (i.e. it is less likely than others) while the mass on \emph{Sales} variable is set to a negligible value. Consequently, the figure on the right side doesn't produce any samples for \emph{Sales} category and only a few samples for \emph{White-Collar} category. Such targeted customizations are difficult to achieve in traditional models.  

\noindent \textbf{Feature Constraints}: We evaluate feasibility by validating whether the counterfactuals satisfy constraints entailed by a given causal model. Similar to ~\cite{mahajan2019preserving}, we consider an unary constraint where it is infeasible for the \emph{Age} variable to decrease in the generated counterfactual and a binary constraint where there is a monotonic trend between the \emph{Age} and \emph{Hours} variables. The former is modeled using a truncated prior while the latter uses the linear approximation in (5). Figure ~\ref{fig_constrposage} highlights that the counterfactual \emph{Age} values are consistently greater than the original when incorporating the constraint through a domain specific prior. Similarly, Figure ~\ref{fig_constragehours} shows that when \emph{Hours} and  \emph{Age} are negatively correlated, modeling their feature dependency explicitly generates counterfactuals that preserve their relationship.

\begin{figure*}[!tbp]
\centering
\includegraphics[width=\textwidth]{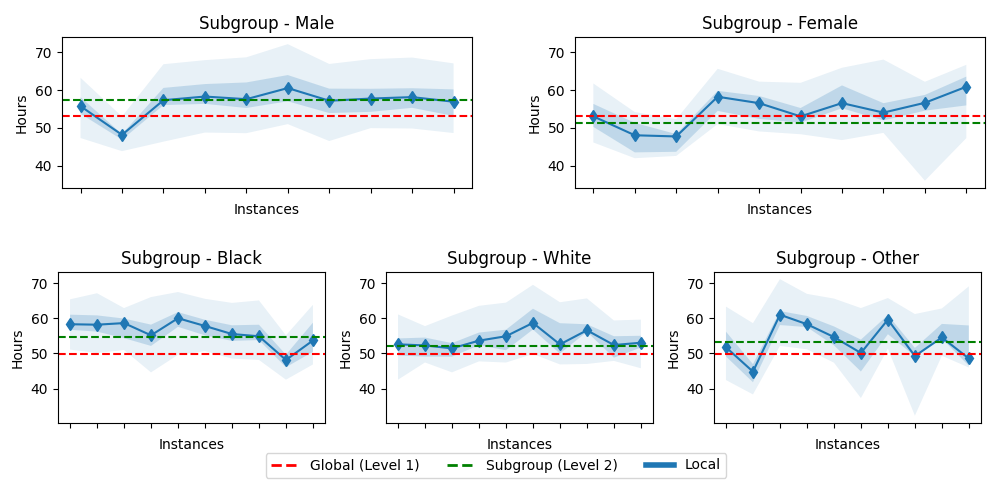}
\caption{Multi-level hierarchical model showing whether the counterfactual of a local instance coincides with or deviates from subgroups within the data (Level 2) and the entire dataset (Level 1) for \emph{Hours} variable in Adult Income. \emph{top}: Subgroups \emph{Male} and \emph{Female} of \emph{Gender} variable. \emph{bottom}: Subgroups \emph{Black}, \emph{White} and \emph{Other} of \emph{Race} variable.}
\label{fig_hier2_ad}
\end{figure*}

\subsection {Multi-level Bayes Evaluation}
Hierarchical models provide the advantage of allowing relative comparisons of counterfactuals generated across different levels in the hierarchy and thus help assess counterfactual fairness. We first consider a two-level hierarchy, and in Figure ~\ref{fig_hier1_ad} compare the values of counterfactual samples generated at the local instance level to the global population level. The left figure uses a box plot to display the median and whiskers of samples from the ordinal \emph{Education} variable in Adult Income dataset. The middle figure plots the mode and dispersion of the nominal \emph{Workclass} variable while the right figure displays the median and credible intervals for the continuous variable \emph{Hours}. The population level aggregated counterfactual values for these variables are shown in a red dashed line. These illustrations allow drawing inferences such as \emph{the education levels required for a local counterfactual seems lower than the global average}, or that \emph{the work category required for local instances is compatible with global standards} and so on. 

For the three-level hierarchy, we focus on the \emph{Hours} variable and consider two different groups corresponding to the \emph{Gender} and \emph{Race} categorical variables. The top part of Figure ~\ref{fig_hier2_ad} plots the median and credible intervals for $Hours$, along with the values at population level (red dashed line) and subgroups level (green dashed line). This enables the simultaneous visualization of how the values at local levels compare against the global values and the \emph{Male} and \emph{Female} subgroups. The bottom part of Figure ~\ref{fig_hier2_ad} contrasts the values for \emph{Hours} against \emph{Black}, \emph{White} and \emph{Other} race categories. See supp. for more results.

\begin{figure*}[!tbp]
\centering
\includegraphics[width=\textwidth]{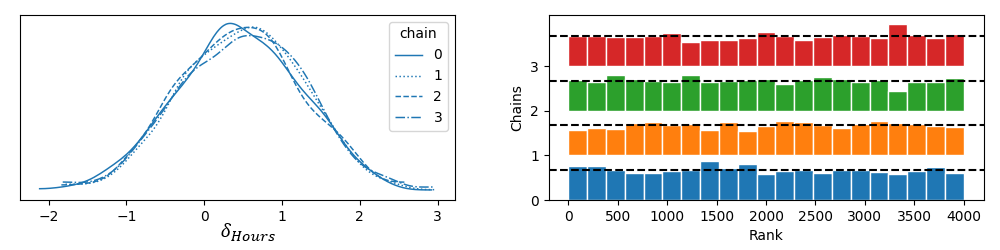}
\caption{Convergence of samples for \emph{Hours} variable in Adult Income dataset. \emph{left}: Density estimate of the posterior samples for four different chains. \emph{right}: Rank plots of posterior draws showing no substantial difference across the chains. } 
\label{fig_convergence}
\end{figure*}

\subsection {Convergence Diagnostics}
It is critical to monitor whether the samples produced using an MCMC method converges ~\cite{roy2020convergence} to the target posterior distribution. The generated samples maybe degenerate if they are strongly correlated with each other and are not effectively independent.  To assess convergence, we run 4 different chains initialized at various starting points and check if the obtained distribution is similar across the chains.  The most widely used convergence diagnostic is the scale reduction factor $\hat{R}$, which compares the variance of all the chains mixed together with the variance of individual chains. The observed $\hat{R}$ was well under $1.1$, thus not detecting any convergence problems. We also inspected  the effective sample size and found it to be large, confirming the absence of auto-correlations in the chains. Figure ~\ref{fig_convergence} displays two other convergence metrics. In the left side, the kernel density estimates of the posterior draws is shown for the \emph{Hours} variable and it can be seen that the distributions appear similar across the chains. Additionally, we visualize the histograms of ranked posterior draws for each chain in the right figure, and as advocated in ~\cite{vehtari2021rank} observe that the rank plots of all chains appear similar thereby indicating a good mixing of the chains.

%% file: sec_conclusion.tex
\section{Limitations}
A valid criticism of the Bayesian methods is the inordinate number of computational steps required for convergence. For collecting $N$ samples of $V$ variables, HMC has a complexity of $\mathcal{O}(NV^{5/4})$. Even though the cost can be amortized for higher levels in the hierarchy and efficient parallelization can scale out the computations, the inference duration may still be intractable for real-time performance. 
The use of HMC also implies that the classifier function must be differentiable. Consequently, our solution is not purely model agnostic. While alternate sampling methods that can handle black box models exist, such solutions may pose challenges in convergence. The advent of AutoGrad and the prevalence of deep learning though makes our differentiable assumption reasonable. 
Finally, in order to model feature dependencies, we assume that the structural causal model is known apriori. However, in practice this information may not be available and learning them automatically is preferable. It must also be noted that we do not support complex causal relationships over multiple features. 

\section {Societal Impact}
Recent studies ~\cite{kasirzadeh2021use,slack2021counterfactual} have highlighted the importance of understanding the vulnerabilities and potential for misuse of counterfactuals. In particular, sufficient attention must be paid to ensure that the generated counterfactuals provide actionable recommendations ~\cite{karimi2021algorithmic}. Our work contributes positively towards allowing people to act, by providing a diverse choice to the users instead of patronizing them with a single ideal option, and personalizing the generation process with informative priors that can offer tailor made solutions. It is also important in a social context to verify whether the interventions vary based on protected attributes~\cite{coston2020counterfactual}. The multi-level setup described here opens a pathway to perform such assessments by comparing the extent of deviations across different subgroups.      

\section{Conclusion}\label{sec_conclusion}
We presented a mechanism to generate multiple alternative counterfactuals in a probabilistic setting and characterized the perturbations at several levels of abstraction. Our formulation can support prior beliefs, handle multimodal parameters, furnish uncertainty metrics and compare across population levels, all while producing a diverse set of choices. The experiment results confirm the benefits offered by the proposed Bayesian framework.  In future, we wish to include complex causal relationships between the features and extend to black box classification models.

%% file: supplement.tex
\section{Evaluation Metrics}
We formally define the quantitative metrics validity, proximity, sparsity and diversity that were used in the experiments here for completeness. Given a test set of negative outcomes $\chi^-$, these metrics are computed as 
\begin{align*}
\begin{split}
    validity & = \frac{|\{\mathbf{x} \in \chi^- \mid \exists\mathbf{x}^*  : f(\mathbf{x}^*) > 0.5\}|}{|\chi^-|} \\
    proximity & = \frac{1}{|\chi^-|}\sum_{\mathbf{x} \in \chi^-}\min_{\mathbf{x}^* \in \mathbf{x}^{cfs}}\left\Vert\mathbf{x}^*-\mathbf{x}\right\Vert_2 \\
    sparsity & = \frac{1}{|\chi^-|}\sum_{\mathbf{x} \in \chi^-}\frac{1}{|\mathbf{x}^{cfs}|}\sum_{\mathbf{x}^* \in \mathbf{x}^{cfs}}\frac{\sum_{l=1}^{d_{cont}} \mathbb{I}(Q(x^*_l),Q(x_l)) + \sum_{l=d_{cont}+1}^{d} \mathbb{I}(x^*_l,x_l)}{d} \\
    diversity(\mathbf{x}^{cfs}) & = \frac{1}{|\mathbf{x}^{cfs}|^2}\sum_{i=1}^{|\mathbf{x}^{cfs}|-1}\sum_{j=i+1}^{|\mathbf{x}^{cfs}|}\Big[\sum_{l=1}^{d_{cont}}\frac{|x^i_l-x^j_l|}{MAD_l}+\sum_{l=d_{cont}+1}^{d} \mathbb{I}(x^i_l,x^j_l)\Big]
\end{split}
\end{align*}
where $\mathbf{x}^{cfs}$ is the set of counterfactual samples of $\mathbf{x}$, $\mathbb{I}$ is an indicator function that evaluates to 1 if both arguments are equal, $Q$ is a quantization function that coarsely bins a continuous feature into 10 discrete intervals and $MAD$ is the median absolute deviation computed from 80\% of the dataset that was marked as training data.

\section{Robustness Evaluation}

We hypothesize that the multi-level Bayes structure enables information sharing through which the quality of parameter estimates can be improved. While there are several themes for quality such as robustness of the estimates w.r.t input perturbations or classification model changes, we focus here on whether using the hierarchical structure enables the counterfactuals to lie in dense regions of the data manifold.  Data supported counterfactuals tend to be model invariant and result in realistic recourses, and hence data density is an important metric for counterfactual robustness.

We consider two measures namely the distance of a counterfactual to its $k$ nearest neighbors and the local outlier factor, which computes the deviation of the local density of a counterfactual with respect to its neighbors. The data points within the positive class in the training dataset is used as the neighborhood.  The distance between any two data points $i$ and $j$ uses the same metric defined above for diversity (the term within the square brackets).

\begin{figure*}[!htbp]
\centering
\includegraphics[width=\linewidth]{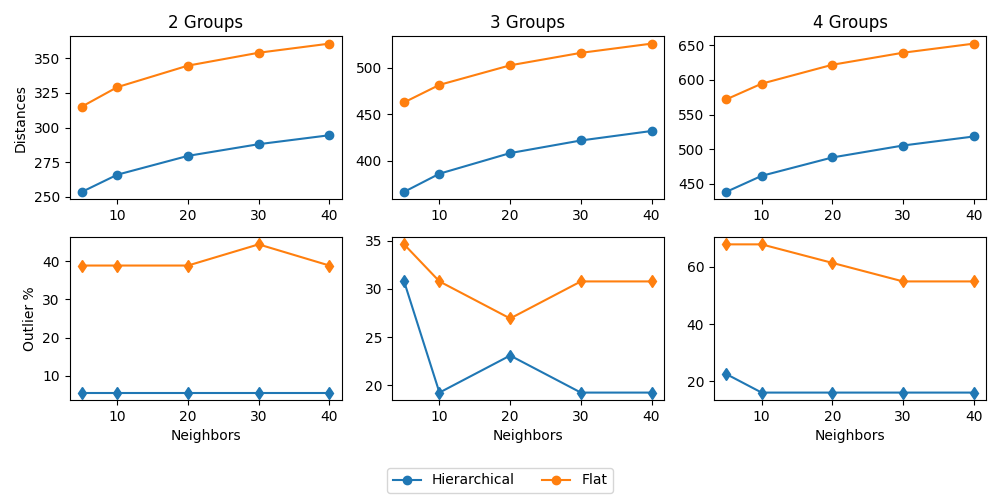}
\caption{Data support for generated counterfactuals in HELOC dataset. \emph{top}: Neighborhood distances. \emph{bottom}: Percentage of outliers. In both cases, the hierarchical model exhibits the desired behavior of a smaller value.} 
\label{fig_robust1}
\end{figure*}

Figure ~\ref{fig_robust1} plots the neighborhood distance and local outlier factor for different values of $k$, from the HELOC dataset. We partition the dataset into different numbers of clusters using the standard k-Means algorithm, and evaluate the generated counterfactual in the neighborhood of its corresponding cluster in the ground-truth. We can see in the top figure that the neighborhood distance is consistently smaller when using a hierarchical model. Similarly, the bottom figure shows that the outlier percentage is reduced when a multi-level Bayes model is utilized. This confirms that using a hierarchical model encourages the desired behavior of a counterfactual to reside in the neighborhood of its subgroup. 

\section{Fairness Evaluation}
Our solution computes the perturbations at different levels of abstraction and it can be exploited to compare the counterfactual of an instance directly with group level counterfactuals. When the groups correspond to protected attributes such as \emph{Gender} or \emph{Race}, it provides a mechanism to verify whether the distributions are identical for each protected group and thereby assess counterfactual fairness.

In Table ~\ref{tab_fairness} we present the cost of recourse both at the subgroup level (left side) and at the instance level (right side) for the AdultIncome dataset. The costs are aggregated using the mean function to handle multiple counterfactual samples. While there maybe many problem specific definitions of recourse cost, we restrict ourselves to the distance function as a proxy for the cost. As before, the distance between an original data point $i$ and its counterfactual $j$ is computed from the $l1$ norm scaled by medium absolute deviation for continuous features and the change in value for categorical features. 

The recourse fairness at a group level is the difference in cost between the subgroups, while at an instance level it is the difference between the cost for a particular instance and the protected group for which we wish to compare against. For example, we can see that the \emph{Race - Other} subgroup has a cost greater than its peers. Similarly, the cost for \emph{Instance 4} is less than both the \emph{Female} and \emph{Black} subgroups the instance belongs to. Given a problem defined scalar threshold for the cost difference, we can now calculate the demographic parity and thus measure fairness.

\begin{table*}[!htbp]
\caption{Recourse cost at the subgroup level (\emph{left}) and at the instances level (\emph{right}) in AdultIncome dataset. The difference in cost can be used to determine recourse fairness. }
\centering
\begin{tabular}{cc}
\begin{tabular}{|l|l|r|}
\hline
Group & Sub Group & Cost \\
\hline
Gender & Male & 5.42 $\pm$ 1.05 \\
Gender & Female & 5.09 $\pm$ 1.01 \\
Race & White & 5.32 $\pm$ 1.02 \\
Race & Black & 5.44 $\pm$ 0.99 \\
Race & Other & 5.53 $\pm$ 1.20 \\
\hline
\end{tabular}
\hspace{1cm}
\begin{tabular}{|l|l|l|l|l|l|r|}
\hline
Instance & Gender & Race & Age & Occupation & Workclass &  Cost \\
\hline
1 & Male & White & 69 & Sales & Self-Emp & 6.74 \\
2 & Male & White & 28 & Blue-Collar & Private & 6.54 \\
3 & Female & White & 52 & White-Collar & Gov & 5.96 \\
4 & Female & Black & 44 & Sales & Private & 4.93 \\
5 & Male & Other & 23 & White-Collar & Private & 7.41 \\
\hline
\end{tabular}

\end{tabular}
\label{tab_fairness}
\end{table*}

\section{Parameter Analysis}
\begin{figure*}[!htbp]
\centering
\includegraphics[width=\linewidth]{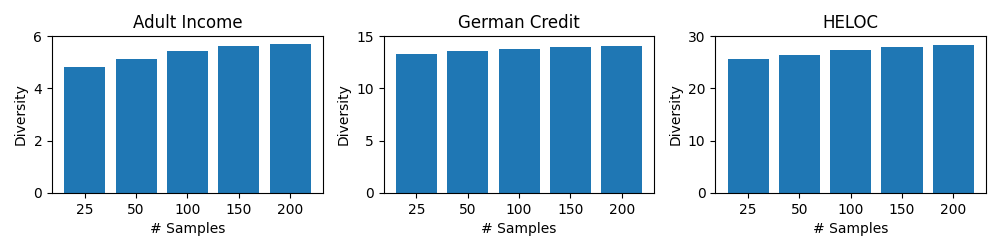}
\caption{Sensitivity analysis on the number of posterior samples. The counterfactual diversity increases with the sample size. } 
\label{fig_paramsens_sample}
\end{figure*}

Besides convergence, the number of posterior samples is also relevant for the diversity in generated counterfactuals. We plot the diversity measure against different numbers of samples in Figure ~\ref{fig_paramsens_sample}. The initial burn-in samples that were used for tuning is ignored and only the samples in equilibrium are considered. We observe that the counterfactual diversity in general increases with the number of samples used across the three datasets. However, they do plateau indicating that only a selected number of truly divergent recourse options is available. In practice, a top-$k$ ranking of these samples based on a domain specific metric maybe necessary before presenting the recommendations to the user.     

\begin{figure*}[!htbp]
\centering
\includegraphics[width=\linewidth]{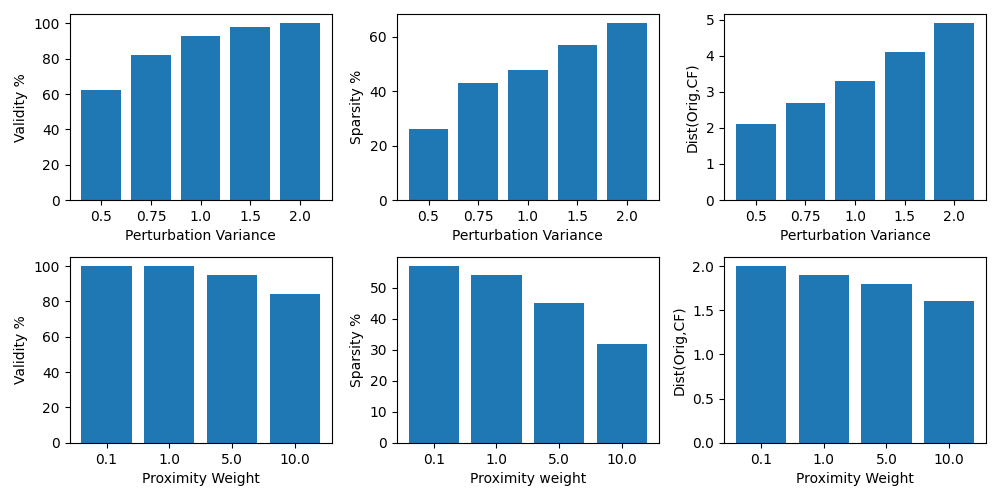}
\caption{Control over the generated counterfactual properties. \emph{top}: Change in the perturbation variance parameter for the HELOC dataset. \emph{bottom}: Modifications to the importance for proximity in Adult Income dataset. } 
\label{fig_paramsens_sigma}
\end{figure*}

The extent of perturbations can be controlled using the prior values. In particular, by adjusting the variance parameter $\sigma$ of a continuous feature's normal distribution, we can influence the generated counterfactuals. The top part of Figure ~\ref{fig_paramsens_sigma} shows how validity, sparsity and proximity changes with $\sigma$ for the HELOC dataset. When the perturbation variance is high, there are more options for constructing the counterfactuals and consequently the number of instances for which a counterfactual can be generated increases (\emph{top left}).  However, it also implies that the percentage of features used (\emph{top center}) and the distance between an original and counterfactual point (\emph{top right}) also increases.  

Modifying the prior parameters for perturbations maybe inconvenient, especially when there is a mixture of categorical and continuous features. An alternate mechanism to vary the generation process is to scale the second term that encourages proximity in equation (11). The bottom part of Figure ~\ref{fig_paramsens_sigma} plots the change in validity, sparsity and proximity for different weights assigned to this term in the Adult Income dataset. As the importance assigned to the proximity between an original and its counterfactual increases, this constrained setting results in the reduction of validity (\emph{bottom left}) and the number of features used (\emph{bottom center}). However, the counterfactual point is now more closer to the original data point (\emph{bottom right}). Both the variance and proximity weight parameters provide effective control to tailor the outcomes based on a problem specific scenario.